\documentclass[letterpaper, 10 pt, conference]{ieeeconf}  

\IEEEoverridecommandlockouts                              
\title{\LARGE \bf
Fly Out The Window: Exploiting Discrete-Time Flatness for Fast Vision-Based Multirotor Flight
}

\author{Melissa Greeff, SiQi Zhou and Angela P. Schoellig\thanks{The authors are with the Dynamic Systems Lab (www.dynsyslab.org) at the University of Toronto Institute for Aerospace Studies (UTIAS) and the Vector Institute for Artificial Intelligence, Toronto. Email: melissa.greeff@mail.utoronto.ca, schoellig@utias.utoronto.ca
This work was supported by Drone Delivery Canada, Defence Research and Development Canada, and Natural Sciences and Engineering Research Council of Canada.}}

\begin{document}

\maketitle
\thispagestyle{empty}
\pagestyle{empty}

\begin{abstract}

Current control design for fast vision-based flight tends to rely on high-rate, high-dimensional and perfect state estimation. This is challenging in real-world environments due to imperfect sensing and state estimation drift and noise. In this letter, we present an alternative control design that bypasses the need for a state estimate by exploiting discrete-time flatness. To the best of our knowledge, this is the first work to demonstrate that discrete-time flatness holds for the Euler discretization of multirotor dynamics. This allows us to design a controller using only a window of input and output information. We highlight in simulation how exploiting this property in control design can provide robustness to noisy output measurements (where estimating higher-order derivatives and the full state can be challenging). Fast vision-based navigation requires high performance flight despite possibly noisy high-rate real-time position estimation. In outdoor experiments, we show the application of discrete-time flatness to vision-based flight at speeds up to 10 m/s and how it can outperform controllers that hinge on accurate state estimation.  

\end{abstract}

\section{INTRODUCTION}

\begin{figure}
     \centering
		\includegraphics[width= 0.45\textwidth, trim={10cm 7cm 2cm 5cm},clip ]{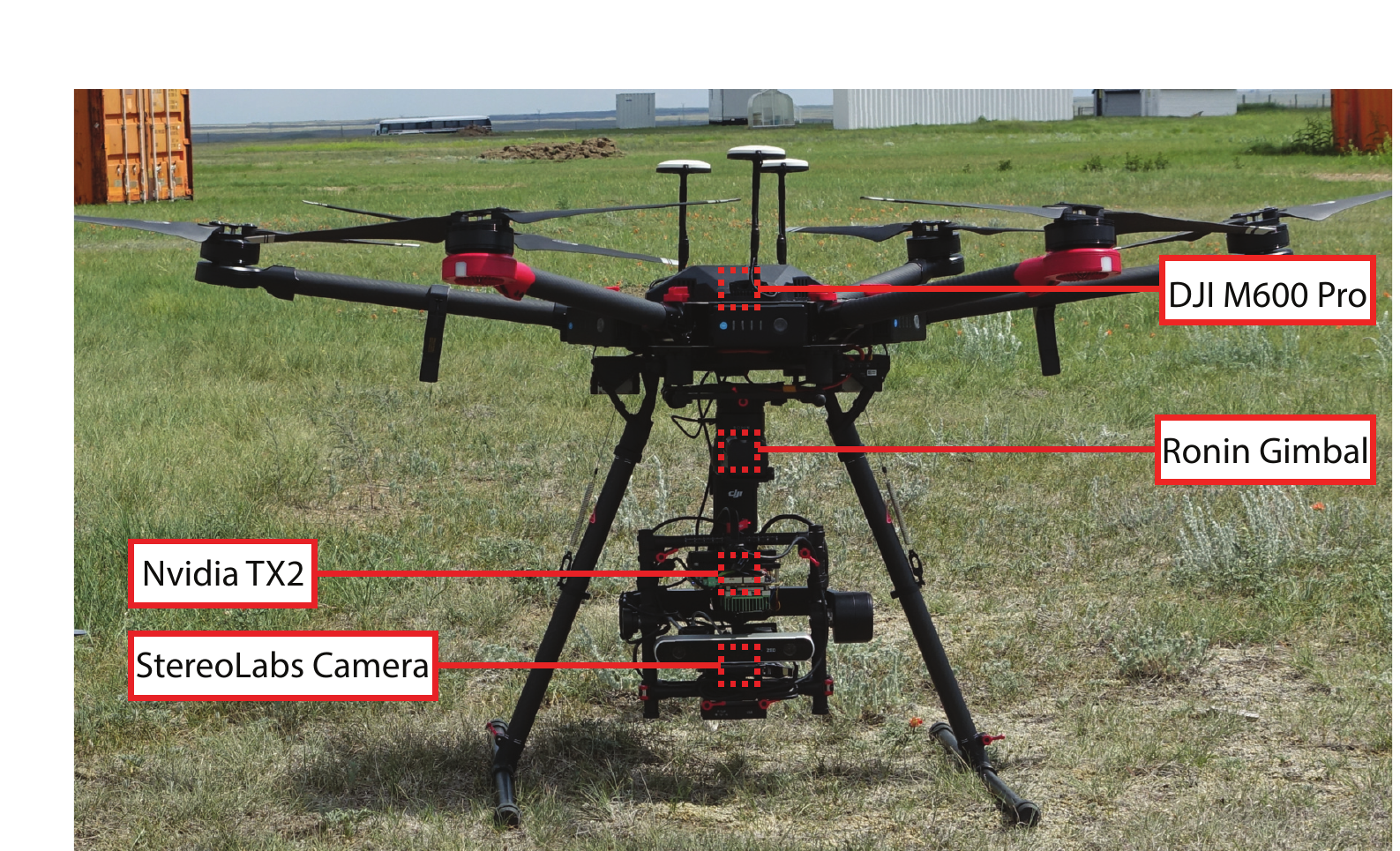}
	      \caption{Multirotor system used for autonomous vision-based flight experiments. We use a DJI M600 Pro with Ronin-MX gimbal and a StereoLabs camera. All computation is performed onboard on the Nvidia TX2.}
	      \label{fig_m600_hardware}
	      \vspace{-0.6cm}
\end{figure}

Multirotor unmanned aerial vehicles (UAVs), see Fig. \ref{fig_m600_hardware}, are mechanically simple and highly maneuverable which makes them suitable to a wide-range of applications such as infrastructure inspection \cite{c5}, search and-rescue missions and mapping operations. This has challenged researchers to develop multirotor systems that can move beyond lab demonstrations to real-world scenarios where GPS or motion capture may not always be available. As such, in recent years significant advancements have been made in enabling faster flight using vision-based sensing as a lightweight and versatile alternative \cite{c20}-\cite{c4}. However, autonomous flight relying on imperfect visual sensing and state estimation drift and noise is still challenging. 

Control design within the visual navigation pipeline tends to rely on high-rate, high-dimensional state estimates (generally 50-200 Hz for position control). In contrast, the visual system often estimates lower-rate (currently 10-35 Hz) and noisy position (and orientation) information. The mismatch between the control specifications and visual measurement requires an additional state estimator to determine high-rate state estimates. Traditionally, the controllers and planners are designed independently of the state estimator \cite{c20}-\cite{c2}. 

Control design tends to assume perfect state estimation. Historically, this has been motivated by the separation principle which theoretically guarantees that optimally is retained for certain linear stochastic systems when optimal control and optimal state estimation are decoupled. However,  the required assumptions for these guarantees, i.e., a linear model and zero-mean Gaussian noise, are practically never satisfied by vision-based multirotor systems in real-world operations. 

Related work tends to address the challenge of high performance control under imperfect state estimation by 1) improving the state estimate by incorporating additional sensors, e.g., IMU, laser range-finders, e.t.c.; 2) designing robust control to account for the worst-case state estimation error learnt offline from data; or 3) performing output feedback control by coupling control and state estimation. In this work, we present an alternative approach that bypasses the state estimation requirement by exploiting a property known as discrete-time flatness. This property allows for control design using a window of inputs and special flat outputs directly. 

\section{RELATED WORK AND CONTRIBUTIONS}

\subsubsection{Improve state estimation} Autonomous vision-based flight typically relies on visual or visual-inertial odometry (VIO), with the addition of an IMU, within a simultaneous localization and mapping (SLAM) \cite{c1} or visual teach and repeat (VT\&R) framework \cite{c20}, \cite{c4}. The challenge of autonomous drone racing is addressed using SLAM \cite{c1}, \cite{c3}. The drift in VIO is compensated for by learning a robust gate (through which the multirotor flies) model and accounting for the static gate locations in an extended Kalman filter (EKF) to estimate the multirotor state. This state can then be used in time-optimal trajectory planners \cite{c2} and simple lower-level controllers, e.g., PD control or model predictive control (MPC). This approach is task-specific and relies on static modelled gates.

\subsubsection{Design robust control for worst-case state estimation} In \cite{c8}, a bound on the state estimation error is learned offline using data. A robust controller is synthesized, using system-level synthesis, that ensures a bounded system response despite estimation errors. This has been applied to vision-based flight for multirotors in \cite{c6}. This results in conservative performance and can sometimes lead to an infeasible problem. Moreover, the current theory applies to linear models and, therefore, has only been applied to flight near hover. 

\subsubsection{Couple control and state estimation} While commonly MPC considers state feedback, output feedback MPC relies on only measurements or outputs. Robust output feedback MPC explicitly considers bounded measurement noise and uses a state estimator within the optimization problem in MPC \cite{c12}. Robust output feedback MPC can be a tube-based method (which guarantees constraint satisfaction and stability despite measurement noise) \cite{c12}-\cite{c11} or a minmax method (which optimizes the worst-case performance) \cite{c10}. Current theory is applied to linear models \cite{c12}-\cite{c11}. Moreover, the main practical limitation is that these methods are slow to compute. Even for low-dimensional linear systems, the controller can often only be implemented at around 5~Hz \cite{c12}. For vision-based navigation, the measurement noise may vary and can be difficult to evaluate prior to flight. Furthermore, overestimating the measurement noise can lead to conservative performance.

In this letter, we present a controller that bypasses the state. Instead our controller relies on only a window of output measurements and previously sent inputs. Our proposed output feedback control hinges on the property of discrete-time flatness. This property allows us to bypass the classical state estimation requirement and work directly with a window of previous inputs and output measurements for feedback control. The contributions of this work are three-fold: 
\begin{itemize}
	\item To the best of our knowledge, this is the first work to demonstrate that the property known as discrete-time flatness holds for the Euler discretization of multirotors. This means that only a window of input (thrust and torques) and output (position and yaw) information is required for control design.
	\item We highlight in simulation how exploiting this property in control design can provide robustness to noisy output measurements (where estimating higher-order derivatives and the full state can be challenging).
	\item Fast vision-based navigation requires high performance flight despite possibly noisy high-rate real-time position estimation. In outdoor experiments, we show the application of discrete-time flatness to vision-based flight at speeds up to 10 m/s and how it can outperform controllers that hinge on accurate state estimation (or higher-order position derivatives). 
\end{itemize}

\section{BACKGROUND}

\subsection{Differential Flatness}
We recall the formal definition of differential flatness \cite{c13}.
\begin{definition}[Differential Flatness]
A nonlinear system model,
\begin{equation}
\mathbf{\dot{x}}(t) = f(\mathbf{x}(t), \mathbf{u}(t))
\label{eq_ct}
\end{equation}
with $t \in \mathbb{R}^+$, $\mathbf{x} \in \mathbb{R}^n$, $\mathbf{u} \in \mathbb{R}^m$ is differentially flat if there exists an output $\mathbf{y}(t) \in \mathbb{R}^m $, whose components are differentially independent (that is, the components are not related to each other through a differential equation), such that the following holds:
$$\mathbf{y} = \tilde{H}(\mathbf{x}, \mathbf{u}, \mathbf{\dot{u}}, ..., \mathbf{u^{\delta}})$$
\vspace{-6mm}
$$\mathbf{x} = \tilde{\Lambda}(\mathbf{y}, \mathbf{\dot{y}}, ..., \mathbf{y}^{r-1})$$
\vspace{-6mm}
\begin{equation}
	\mathbf{u} = \tilde{F}(\mathbf{y}, \mathbf{\dot{y}}, ..., \mathbf{y}^r)
	\label{eq_diff_flat_input}
\end{equation}
where $\tilde{H}, \tilde{\Lambda}, \tilde{F}$ are smooth functions, $\delta$ and $r$ are the
maximum orders of the derivatives of $\mathbf{u}$ and $\mathbf{y}$ needed to
describe the system and $\mathbf{y} = [y_1, y_2, ...y_m]^T$ is called the flat output. $\square$
\label{def_diff_flat}
\end{definition}

More intuitively, the system (\ref{eq_ct}) is flat if there exists a one-to-one correspondence (or mapping) between its solutions $(\mathbf{x}(t), \mathbf{u}(t))$ and solutions $\mathbf{y}(t)$ of a trivial systems (with the same number of inputs, i.e., $\mathbf{y}^r = \mathbf{v}$). This means that both the state $\mathbf{x}(t)$ and the input $\mathbf{u}(t)$ at time $t$ can be determined from the (flat) output $\mathbf{y}(t)$ and a finite number of its derivatives. 

The property of differential flatness (with position and yaw as output) is well-established for multirotors \cite{c14} and has been exploited for efficient trajectory generation in \cite{c14} and model predictive control in \cite{c15}. 

\subsection{Discrete-Time Flatness}
The discrete-time counterpart of differential flatness, commonly referred to as forward-difference flatness (or difference flatness), has recently been introduced in \cite{c16}\cite{c17}. Roughly speaking, the main difference is that time derivatives in Definition~\ref{def_diff_flat} are replaced by forward-shifts of the output \cite{c18}.

\begin{definition}[Discrete-Time Flatness]
A nonlinear system model,
\begin{equation}
	\mathbf{x}_{k+1} = f(\mathbf{x}_k, \mathbf{u}_k)
	\label{eq_df}
\end{equation}
with $k \geq 0$, $\mathbf{x}_k \in \mathbb{R}^n$, $\mathbf{u}_k \in \mathbb{R}^m$ is difference flat if there exists an output $\mathbf{y}_k \in \mathbb{R}^m $, whose components are independent (that is, the components are not related to each other through a difference equation), such that the following holds:
$$ \mathbf{y}_k = H(\mathbf{x}_k, \mathbf{u}_k, \mathbf{u}_{k+1}, ..., \mathbf{u}_{k+\delta})$$
\begin{equation}
	\mathbf{x}_k = \Lambda \left(\mathbf{y}_k, \mathbf{y}_{k+1}, ..., \mathbf{y}_{k+r-1} \right)
	\label{eq_df_state}
\end{equation}
\begin{equation}
	\mathbf{u}_k = F \left(\mathbf{y}_k, \mathbf{y}_{k+1}, ..., \mathbf{y}_{k+r} \right)
	\label{eq_df_input}
\end{equation}
where $H, \Lambda, F$ are smooth functions, $\delta$ and $r$ are the
maximum number of forward shifts of $\mathbf{u}$ and $\mathbf{y}$ needed to
describe the system and $\mathbf{y}_k = [y_{1,k}, y_{2,k}, ...y_{m,k}]^T$ is called the flat output. $\square$
\label{def_dis_flat}
\end{definition}

More intuitively, the system (\ref{eq_df}) is flat if there exists a one-to-one correspondence (or mapping) between its solutions $(\mathbf{x}_k, \mathbf{u}_k)$ and solutions $\mathbf{y}_k$ of a trivial system (i.e., they do not satisfy a difference equation, but rather $\mathbf{y}_{k+r} = \mathbf{v}_k$). This means that both the state $\mathbf{x}_k$ and input $\mathbf{u}_k$ at time step $k$ can be determined from a finite number of future values of the output $\mathbf{y}_k$ \cite{c18}. The key concept of discrete-time flatness is that \textit{the trajectory of the (flat) output in a certain finite window uniquely determines the state and input at any time step} \cite{c19}.

Unfortunately, differential flatness of (\ref{eq_ct}) does not necessarily imply difference flatness of a discretization (either exactly if possible or using Euler discretization) of (\ref{eq_ct}). A counterexample is provided in \cite{c18}. However, in this paper we show that difference flatness still applies to an Euler discretization of the multirotor dynamics and how we can exploit this property in \textit{control design using only input and output information} (i.e., bypassing the more traditional requirement for a higher-dimensional state). 

\section{METHODOLOGY}
We show that discrete-time flatness, as described in Definition~\ref{def_dis_flat}, holds for the Euler discretization of the full multirotor dynamics model (from \cite{c13}). We also, briefly, show that this property holds for a simpler 2-D motion model, with underlying first-order pitch and roll dynamics (as a result of lower-level attitude controllers), that is used in our simulation and experimental results in Sec. \ref{sec_sim} - \ref{sec_exp}. In Sec. \ref{meth_c}, we illustrate how discrete-time flatness can be used in predictive control design that uses only input and (flat) output information.  

\subsection{Discrete Flatness of Euler-Discretized Multirotor Dynamics Model}
\label{meth_a}

\subsubsection{Dynamics Model}

We consider the following discrete-time model (using the Euler discretization approximation of the model in \cite{c13}):
\begin{equation*}
	\mathbf{x}_{k+1} = \mathbf{x}_{k} + \delta t \mathbf{\dot{x}}_k, 
\end{equation*}
where $\delta t$ is the time step of the discretization and the state at time step $k$, $\mathbf{x}_{k} = [x_k, \dot{x}_k, y_k, \dot{y}_k, z_k, \dot{z}_k,  \theta_k,  \phi_k, \psi_k, p_k, q_k, r_k]^T$ comprises of the 3-D position $x_k, y_k, z_k$, the 3-D velocity $\dot{x}_k, \dot{y}_k, \dot{z}_k$, the vehicle roll $\phi_k$, pitch  $\theta_k$ and yaw $\psi_k$, and the angular velocities (in the vehicle frame) $p_k, q_k, r_k$ at time step $k$. The translational dynamics are:
\begin{equation}
	\begin{bmatrix}
		\ddot{x}_k \\ \ddot{y}_k \\ \ddot{z}_k + g
	\end{bmatrix} = \frac{T_k}{m} \begin{bmatrix}
		R^{13}_k \\ R^{23}_k \\ R^{33}_k
	\end{bmatrix},
	\label{eq_dyn_trans}
\end{equation} 
where $\ddot{x}_k,\ddot{y}_k,\ddot{z}_k$ are the 3-D accelerations at time step $k$, $T_k$ is the commanded thrust at time step $k$, $g$ is the gravitational constant, $m$ is the vehicle mass, and $R$ is the rotation matrix from the body to inertial frames. The notation $R^{\alpha \beta}_k$ denotes the $\alpha$ row and $\beta$ column of rotation matrix $R$ at time step $k$. The rotational dynamics are:
\begin{equation}
	\begin{bmatrix}
		\dot{p}_k \\ \dot{q}_k \\ \dot{r}_k
	\end{bmatrix} = \begin{bmatrix} -\frac{(I_{zz} - I_{yy})}{I_{xx}} q_k r_k + \frac{1}{I_{xx}} \tau^{x}_k \\ -\frac{(I_{xx} - I_{zz})}{I_{yy}} p_k r_k + \frac{1}{I_{yy}} \tau^{y}_k \\  -\frac{(I_{yy} - I_{xx})}{I_{zz}} p_k q_k + \frac{1}{I_{zz}} \tau^{z}_k \end{bmatrix},
	\label{eq_dyn_rot}
\end{equation} 
where $\dot{p}_k,\dot{q}_k,\dot{r}_k$ are the angular accelerations and $\tau^{x}_k, \tau^{y}_k, \tau^{z}_k$ are the commanded torques, about the respective axes, at time step $k$. We have assumed a diagonal inertial matrix with diagonal components $I_{xx}, I_{yy}, I_{zz}$.

 \subsubsection{Discrete-Time Flatness Derivation}
 
 We consider the output comprising of the 3-D position and yaw at time step $k$: 
 $$\mathbf{y}_k = [x_k, y_k, z_k, \psi_k]^T. $$
 We show that both state $\mathbf{x}_k$, (\ref{eq_df_state}) in Definition \ref{def_dis_flat}, and input $\mathbf{u}_k = [T_k, \tau^{x}_k, \tau^{y}_k, \tau^{z}_k]^T$, (\ref{eq_df_input}) in Definition \ref{def_dis_flat}, can be determined from forward shifts of the output $\mathbf{y}_k$.
 
 \paragraph{State from Forward Shifts of Output} From the discretized position dynamics we obtain the velocity at time step $k$ in terms of forward shifts of the output, i.e., $\dot{x}_k = \frac{x_{k+1} - x_{k}}{\delta t}$ and $\dot{y}_k = \frac{y_{k+1} - y_{k}}{\delta t}$, and $\dot{z}_k = \frac{z_{k+1} - z_{k}}{\delta t}$. Similarly, velocity at time step $k+1$ can be described in terms of forward shifts of the output, i.e., $\dot{x}_{k+1} = \frac{x_{k+2} - x_{k+1}}{\delta t}$, $\dot{y}_{k+1} = \frac{y_{k+2} - y_{k+1}}{\delta t}$, $\dot{z}_{k+1} = \frac{z_{k+2} - z_{k+1}}{\delta t}$. Given that $[R^{13}_k, R^{23}_k, R^{33}_k]^T$ is a unit vector in (\ref{eq_dyn_trans}), we can obtain this vector from forward shifts of the output as $[R^{13}_k, R^{23}_k, R^{33}_k]^T = \frac{\mathbf{t}_k}{|\mathbf{t}_k|}$ where:
 \begin{equation}
 	\mathbf{t}_k = \begin{bmatrix}
 		\frac{x_{k+2} - 2 x_{k+1} + x_k}{\delta t^2} \\ \frac{y_{k+2} - 2 y_{k+1} + y_k}{\delta t^2} \\ \frac{z_{k+2} - 2 z_{k+1} + z_k}{\delta t^2} + g
 	\end{bmatrix}.
 	\label{eq_t}
 \end{equation}
We obtain the pitch $\theta_k$ and roll $\phi_k$ at time step $k$ in terms of 2 forward shifts of the output by plugging in the above expressions for $R^{13}_k, R^{23}_k, R^{33}_k$ into:
\begin{align}
	\theta_k &= atan \left(\frac{R^{13}_k}{R^{33}_k} C_{\psi_k} + \frac{R^{23}_k}{R^{33}_k} S_{\psi_k}  \right) , \label{eq_pitch} \\
	\phi_k &= atan \left( \left(\frac{R^{13}_k}{R^{33}_k} S_{\psi_k} - \frac{R^{23}_k}{R^{33}_k} C_{\psi_k}\right) C_{\theta_k} \right),
	\label{eq_roll}
\end{align}
where we have assumed $\theta_k \in (-\pi/2, \pi/2)$ and $\phi_k \in (-\pi/2, \pi/2)$. From these Euler angles, we can describe the rotation matrix $R_k$ at time step $k$ in terms of forward shifts of the output. We can obtain similar expressions for $R^{13}_{k+1}, R^{23}_{k+1}, R^{33}_{k+1}$ at time step $k+1$ by computing $\mathbf{t}_{k+1}$ in (\ref{eq_t}). We determine the pitch $\theta_{k+1}$ and roll $\phi_{k+1}$ at time step $k+1$ by using $R^{13}_{k+1}, R^{23}_{k+1}, R^{33}_{k+1}$ and $\psi_{k+1}$ in (\ref{eq_pitch}) - (\ref{eq_roll}). 
We can obtain similar expressions for the pitch $\theta_{k+2}$ and roll $\phi_{k+2}$ at time step $k+2$. The angular velocity of the vehicle in the inertial frame $\omega_k$ at time step $k$ is (see \cite{c13} for details):
\begin{equation*}
	\omega_k = \begin{bmatrix}
		C_{\psi_k} & R^{12}_k & 0 \\ S_{\psi_k} & R^{22}_k & 0 \\ 0 & R^{32}_k & 1
	\end{bmatrix} \begin{bmatrix}
		\dot{\phi}_k \\ \dot{\theta}_k \\ \dot{\psi}_k
	\end{bmatrix} = R_k \begin{bmatrix}
		p_k \\ q_k \\ r_k
	\end{bmatrix}.
\end{equation*}
Exploiting the Euler discretization, we obtain:
\begin{equation}
\begin{bmatrix}
	p_k \\ q_k \\ r_k
\end{bmatrix} = R^{-1}_k \begin{bmatrix}
		C_{\psi_k} & R^{12}_k & 0 \\ S_{\psi_k} & R^{22}_k & 0 \\ 0 & R^{32}_k & 1
	\end{bmatrix} \begin{bmatrix}
		\frac{\phi_{k+1} - \phi_k}{dt} \\ \frac{\theta_{k+1} - \theta_k}{dt} \\ \frac{\psi_{k+1} - \psi_k}{dt}
	\end{bmatrix}. 
	\label{eq_ang_vel}
\end{equation}
Using the relationship between forward shifts of the output $\mathbf{y}_k$ and the Euler angles $\theta_k, \phi_k, \psi_k$ at time step $k$, see (\ref{eq_pitch}) - (\ref{eq_roll}), and $\theta_{k+1}, \phi_{k+1}, \psi_{k+1}$ at time step $k+1$ in (\ref{eq_ang_vel}), we can determine the angular velocities $p_k, q_k, r_k$ at time step $k$ in terms of forward shifts of the output. Similarly we can obtain expressions for $p_{k+1}, q_{k+1}, r_{k+1}$ by plugging in the expressions for $\phi_{k+1}, \theta_{k+1}, \psi_{k+1}$ and $\phi_{k+2}, \theta_{k+2}, \psi_{k+2}$ in terms of the forward shifts of the output. 

\paragraph{Input from Forward Shifts of Output} We can obtain the commanded thrust at time step $k$ in terms of forward shifts of the output from (\ref{eq_dyn_trans}) as $T_k = m |\mathbf{t}_k|$, where $|\mathbf{t}_k|$ is the magnitude of $\mathbf{t}_k$ in (\ref{eq_t}). Using the Euler discretization of the rotational dynamics (\ref{eq_dyn_rot}), the torques at time step $k$ are:
 \begin{equation}
 	\begin{bmatrix}
 		\tau^{x}_k \\ \tau^{y}_k \\ \tau^{z}_k 
 	\end{bmatrix} = \begin{bmatrix}
 		I_{xx} \frac{p_{k+1} - p_k}{dt} + \left(I_{zz} - I_{yy}\right) q_k r_k \\ 
 		I_{yy} \frac{q_{k+1} - q_k}{dt} + \left(I_{xx} - I_{zz}\right) p_k r_k \\
 		I_{zz} \frac{r_{k+1} - r_k}{dt} + \left(I_{yy} - I_{xx}\right) p_k q_k \\
 	\end{bmatrix}.
 	\label{eq_torq}
 \end{equation}
Using the expressions for angular velocities $p_k, q_k, r_k$ at time step $k$ and $p_{k+1}, q_{k+1}, r_{k+1}$ at $k+1$ in terms of the forward shifts of the output allows us to compute the torques in (\ref{eq_torq}) from 4 forward shifts of the output (i.e., we require $\mathbf{y}_k, \mathbf{y}_{k+1}, \mathbf{y}_{k+2}, \mathbf{y}_{k+3}, \mathbf{y}_{k+4}$). From (\ref{eq_df_input}) in Definition \ref{def_dis_flat}, we call this a window of $r=4$. 
 
\subsection{Discrete Flatness of Multirotor Model in 2-D with First-Order Pitch-Roll Dynamics}
\label{meth_b}

We consider a slightly simpler model in the simulations, Sec. \ref{sec_sim}, and experiments, Sec. \ref{sec_exp}, performed in this letter. The multirotor performs only 2-D motion in the $x-y$ plane and we have an underlying attitude controller. The underlying attitude controller generates a first-order pitch and roll response. This is a common (and useful) assumption for many commercial platforms (such as the DJI M600 in Fig. \ref{fig_m600_hardware}) where a black-box attitude controller is  implemented onboard \cite{c15}. 

\subsubsection{Dynamics Model}
We consider the following discrete-time model (using Euler discretization approximation):
\begin{equation*}
	\mathbf{x}_{k+1} = \mathbf{x}_{k} + \delta t \mathbf{\dot{x}}_k, 
\end{equation*}
where the state at time step $k$, $\mathbf{x}_{k} = [x_k, \dot{x}_k, y_k, \dot{y}_k, \theta_k, \phi_k]^T$ comprises of  the 2-D position $x_k$, $y_k$, the 2-D velocity $\dot{x}_k$, $\dot{y}_k$, and the pitch $\theta_k$ and roll $\phi_k$ angles. We consider the dynamics:
\begin{equation}
	\mathbf{\dot{x}}_{k} = [\dot{x}_k, g \frac{R^{13}_k}{R^{33}_k}, \dot{y}_k, g \frac{R^{23}_k}{R^{33}_k}, \frac{\tilde{k}}{\tau} \theta^{cmd}_k -\frac{\theta_k}{\tau}, \frac{\tilde{k}}{\tau} \phi^{cmd}_k - \frac{\phi_k}{\tau}]^T,
	\label{eq_dyn_2d}
\end{equation}
where $R$ is the rotation matrix from the body to inertial frames, $R^{\alpha \beta}_k$ is the $\alpha$ row and $\beta$ column of rotation matrix $R$ at time step $k$, $\tau$ is the first-order time constant and $\tilde{k}$ is the first-order gain for the roll and pitch dynamics. The control input $\mathbf{u}_{k} = [\theta^{cmd}_k, \phi^{cmd}_k]^T$ includes the commanded pitch and roll. We assume the yaw is fixed, i.e., $\psi_k = \psi_{\star} \forall k$.

 \subsubsection{Discrete-Time Flatness Derivation}
 
 We consider the output comprising of the position at time step $k$: 
 $$\mathbf{y}_k = [x_k, y_k]^T. $$
 We show that both state $\mathbf{x}_k$ and input $\mathbf{u}_k$ can be determined from forward shifts of the output $\mathbf{y}_k$.
  
 \paragraph{State from Forward Shifts of Output} Similar to Sec. \ref{meth_a}, we can obtain the 2-D velocity at time step $k$ and $k+1$ in terms of forward shifts of the output.  Similarly, by using the 2-D velocity at time step $k$ and $k+1$ in (\ref{eq_dyn_2d}), we obtain the rotation at time step $k$ in terms of forward-shifts of the output. That is, $\frac{R^{13}_k}{R^{33}_k} = \frac{x_{k+2} - 2 x_{k+1} + x_k}{g \delta t^2} $ and $\frac{R^{23}_k}{R^{33}_k} = \frac{y_{k+2} - 2 y_{k+1} + y_k}{g \delta t^2}$. The pitch $\theta_k$ and roll $\phi_k$ at time step $k$ in terms of forward shifts of the output is obtain from (\ref{eq_pitch}) - (\ref{eq_roll}) with $\psi_k = \psi_{\star}$.

 \paragraph{Input from Forward Shifts of Output} Similar to Sec. \ref{meth_a}, the rotation at time step $k+1$, i.e., $\frac{R^{13}_{k+1}}{R^{33}_{k+1}} = \frac{x_{k+3} - 2 x_{k+2} + x_{k+1}}{g dt^2}$ and $\frac{R^{23}_{k+1}}{R^{33}_{k+1}} = \frac{y_{k+3} - 2 y_{k+2} + y_{k+1}}{g dt^2}$, is used to determine the pitch $\theta_{k+1}$ and roll $\phi_{k+1}$ at $k+1$ in terms of forward shifts of the output. From the first-order pitch and roll dynamics we can determine the pitch and roll commands $\theta^{cmd}_k$ and $\phi^{cmd}_k$ from the pitch and roll at time steps $k$ and $k+1$, i.e.:
 \begin{align*}
 	\theta^{cmd}_k &= \frac{1}{\tilde{k}} \left(\tau \theta_{k+1} - \theta_k \right), \quad 
 	\phi^{cmd}_k &= \frac{1}{\tilde{k}} \left(\tau \phi_{k+1} - \phi_k \right).
 \end{align*}
The pitch and roll at time steps $k$ and $k+1$ can be determined from 3 forward shifts of the output. Consequently, we have determined the input at time step $k$ as a function of $3$ forward shifts of the output, i.e.,
\begin{equation}
	\mathbf{u}_k = F\left(\mathbf{y}_k, \mathbf{y}_{k+1}, \mathbf{y}_{k+2}, \mathbf{y}_{k+3}\right).
	\label{eq_outin}
\end{equation}
We coin this function (\ref{eq_outin}) the \textit{Output-To-Input Map} with a window $r=3$ in (\ref{eq_df_input}). The \textit{Input-To-Output Map} is described by its inverse, i.e.,
\begin{equation}
	\mathbf{y}_{k+3} = F^{-1}\left(\mathbf{y}_k, \mathbf{y}_{k+1}, \mathbf{y}_{k+2}, \mathbf{u}_k\right).
	\label{eq_inout}
\end{equation}

\subsection{Predictive Control Exploiting Discrete Flatness}
\label{meth_c}

\begin{figure}
     \centering
		\includegraphics[width= 0.48\textwidth]{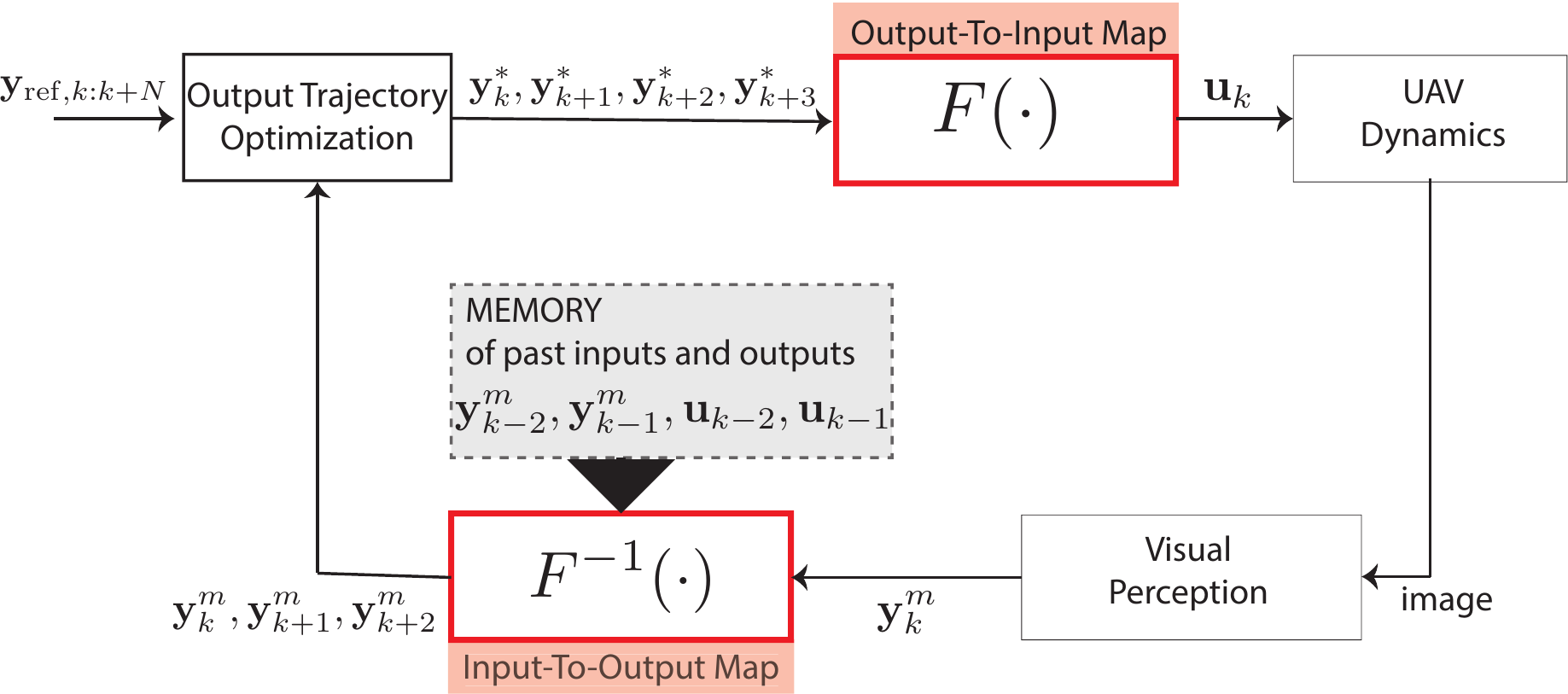}
	      \caption{Overview of predictive control (horizon $N$) design using a window $r=3$ of input and (flat) output data by exploiting discrete-time flatness.}
	      \label{fig_arch_diag}
	      \vspace{-0.6cm}
\end{figure}

We present a predictive control design in Fig. \ref{fig_arch_diag} that exploits discrete-time flatness, from Definition \ref{def_dis_flat}, to use a window of previous inputs and measured (flat) outputs to compute the control input. To do this, we propose three core components: \textit{1) Output-to-Input Map} - maps a window $r$ of the future optimized trajectory to the input using (\ref{eq_output_to_input}); \textit{2) Input-To-Output Map} - maps a window of past inputs and output measurements to constraint the future trajectory using (\ref{eq_input_to_output}); and \textit{3) Output Trajectory Optimization} - optimizes the output trajectory using (\ref{eq_optimization}). 
\subsubsection{Output-To-Input Map}
At time step $k$ we require a window of future time steps of the output trajectory $\mathbf{y}_k, {\mathbf{y}_{k+1}}, ... {\mathbf{y}_{k+r}}$ to determine the input $\mathbf{u}_k$. We propose using predictive control to predict the optimized output trajectory window $\mathbf{y}^*_k, {\mathbf{y}^*_{k+1}}, ... {\mathbf{y}^*_{k+r}}$ which can be used to determined the input from (\ref{eq_df_input}) as:
\begin{equation}
	\mathbf{u}_k = F\left(\mathbf{y}^*_k, {\mathbf{y}^*_{k+1}}, ... {\mathbf{y}^*_{k+r}}\right).
	\label{eq_output_to_input}
\end{equation}
For example, we use the optimized output trajectory window $\mathbf{y}^*_k, {\mathbf{y}^*_{k+1}}, ... {\mathbf{y}^*_{k+3}}$ in the \textit{Output-to-Input Map} in (\ref{eq_outin}) for our 2-D multirotor model (Sec. \ref{meth_b}) with a window size $r=3$.

\subsubsection{Input-To-Output Map}
At time step $k$ we require a window of past time step measurements of the output trajectory $\mathbf{y}^m_k, \mathbf{y}^m_{k-1},\mathbf{y}^m_{k-r+1}$ and previously  sent inputs $\mathbf{u}_{k-1}, ...,\mathbf{u}_{k-r+1}$ to determine the effect of past inputs on the future output trajectory at $\mathbf{y}^m_{k+1}, ...\mathbf{y}^m_{k+r-1}$. More precisely, we compute $\mathbf{y}^m_{k+1}, ..., \mathbf{y}^m_{k+r-1}$ as:
\begin{equation}
	\mathbf{y}^m_{k+(j-1)} = F^{-1}\left(\mathbf{y}^m_{k-(r-j+1)},..., \mathbf{y}^m_{k+(j-2)}, \mathbf{u}_{k-(r-j+1)}\right),
	\label{eq_input_to_output}
\end{equation}
where $j=2,...,r$ and $r$ is the window size. For example, the \textit{Input-to-Output Map} in (\ref{eq_inout}) for our 2-D multirotor model (Sec. \ref{meth_b}) determined a window size $r=3$. In this case, we compute the effect of $u_{k-1}$ and $u_{k-2}$ on $y^m_{k+1}$ and $y^m_{k+2}$ as:
\vspace{-0.4cm}
\begin{align*}
	\mathbf{y}^m_{k+1} &= F^{-1}\left(\mathbf{y}^m_{k-2}, \mathbf{y}^m_{k-1}, \mathbf{y}^m_{k}, \mathbf{u}_{k-2}\right), \\
	\mathbf{y}^m_{k+2} &= F^{-1}\left(\mathbf{y}^m_{k-1}, \mathbf{y}^m_{k}, \mathbf{y}^m_{k+1}, \mathbf{u}_{k-1}\right).
\end{align*}
This is used to constrain the output trajectory optimization to account for the effects of previously sent inputs. 

\subsubsection{Output Trajectory Optimization}
At each time step, we solve the following optimization problem:
\begin{equation}
\begin{aligned}
\mathbf{y}^*_{k:k+N} = \underset{\mathbf{y}_{k:k+N}}{\operatorname{argmin}} \quad & J\left(\mathbf{y}_{k:k+N}\right) \\
\textrm{s.t.} \quad & \mathbf{y}_{k+j-1} = \mathbf{y}^{m}_{k+j-1}, \forall j=1,..,r\\
\end{aligned}
\label{eq_optimization}
\end{equation}
where $N$ is the prediction horizon, $J(\cdot)$ is a selected trajectory cost function (often selected as a quadratic with a tracking error term and a regularization term regulating how quickly the output changes). Additional constraints on the output trajectory can also be enforced if necessary.  

\section{SIMULATIONS}
\label{sec_sim}

\begin{figure}
     \centering
		\includegraphics[width= 0.5\textwidth]{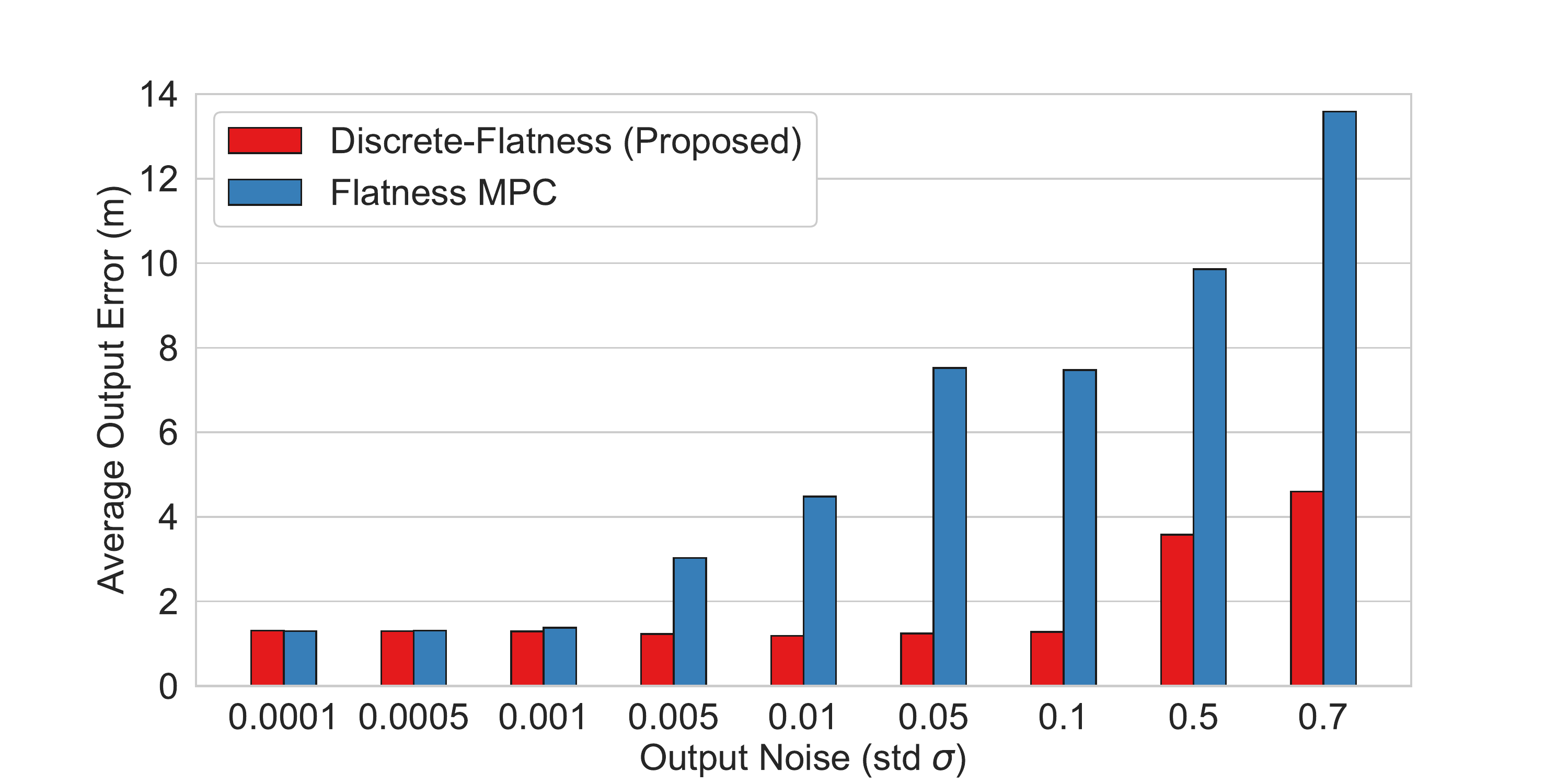}
	      \caption{Comparison of the effect of output noise on the performance of predictive control exploiting continuous-time flatness (\textit{Flatness MPC} in blue) and discrete-time flatness (\textit{Discrete-Flatness} in red). Output measurements of the $x$-position have zero-mean gaussian noise $\mathbf{y}^{m} = \mathbf{y} + [\mathcal{N}(0, \sigma^2), 0]^T$ where we varying the standard deviation $\sigma$ (Output Noise). For each $\sigma$, we compute the average output error (for a fixed reference $\mathbf{y}_{ref} = [10, 0]^T $) over 10 trials. The proposed \textit{Discrete-Flatness} (red) is more robust to output noise compared to the traditional \textit{Flatness MPC} (blue) which relies on higher-order position derivatives.}
	      \label{fig_simcom_noise}
	      \vspace{-0.6cm}
\end{figure}

In these simulation, we consider the dynamics in (\ref{eq_dyn_2d}) from Sec. \ref{meth_b} and consider 1-D motion in the $x$-direction with $\psi^* = 0$, $y = 0$, $\dot{y} = 0$ and $\phi = 0$. We consider a fixed reference $\mathbf{y}_{ref,k} = [10, 0]^T \quad \forall k$. We start from no motion at the origin, i.e., $\mathbf{x}_k= \mathbf{0}$, $\mathbf{y} = \mathbf{0}$ at $k=0$. The dynamics (\ref{eq_dyn_2d}) are executed at 200~Hz. The controllers are run at 50~Hz. 

\paragraph{Discrete-Flatness} We consider the proposed predictive control exploiting discrete-time flatness with the cost in (\ref{eq_optimization}):
\vspace{-0.2cm}
\begin{equation*}
	J(\cdot) = \sum_{k=0}^{N}  \left(\mathbf{y}_k - \mathbf{y}_{ref,k}\right)^T \mathbf{Q} \left(\mathbf{y}_k - \mathbf{y}_{ref,k}\right)  + \mathbf{v}_k^T \mathbf{R} \mathbf{v}_k
	\vspace{-0.1cm}
\end{equation*}
where $\mathbf{v}_k =\mathbf{y}_{k+3} - 3 \mathbf{y}_{k+2} + 3 \mathbf{y}_{k+1} - \mathbf{y}_k$, $N=200$ is the prediction horizon, $\mathbf{Q}$ weights the error with the reference point and $\mathbf{R}$ weights a regularisation term. 

\paragraph{Flatness MPC (FMPC)} We consider a MPC that exploit differential flatness (continuous) as in \cite{c15}. While this is a continuous-time property, practically the controller is  implemented in discrete-time. In FMPC, we consider the discretized version of the linear models at 50~Hz $\dddot{y} = v_1$ and $\dddot{y} = v_2$ (integrator chains of degree 3) where the flat inputs are $v_1$, $v_2$. The flat state $\mathbf{z} = [x, \dot{x}, \ddot{x}, y, \dot{y}, \ddot{y}]^T$ and the flat input $\tilde{\mathbf{v}} = [v_1, v_2] = [\dddot{x}, \dddot{y}]$. We optimize for the flat input (or jerk) $\tilde{\mathbf{v}}_k$ at time step $k$ in MPC and consider a cost similar to predictive control using discrete-time flatness:
\vspace{-0.1cm}
\begin{equation*}
	J(\cdot) = \sum_{k=0}^{N}  \left(\mathbf{y}_k - \mathbf{y}_{ref,k} \right)^T \mathbf{Q} \left(\mathbf{y}_k - \mathbf{y}_{ref,k}\right)  + \tilde{\mathbf{v}}_k^T \mathbf{\tilde{R}} \tilde{\mathbf{v}}_k
	\vspace{-0.1cm}
\end{equation*}
where $\mathbf{Q}$ weights the tracking error, $\mathbf{\tilde{R}}$ weights the jerk (third derivative of position) and $N=200$ is the prediction horizon. The input $\mathbf{u}_k$ is computed using feedback linearization, i.e., using (\ref{eq_diff_flat_input}) with flat state $\mathbf{z}_k$ and the optimized $\mathbf{v}_k$. The main difference is that this approach relies on a state requiring higher-order derivatives of the output (or position). These higher-order derivatives (i.e., velocity and acceleration) are computed as first-order finite differences of the measured output. 

\paragraph{Noisy Output Measurements}
We consider noisy output measurements of the $x$-position with zero-mean Gaussian noise (and no motion in $y$):
\begin{equation*}
	\mathbf{y}^{m} = \mathbf{y} + [\mathcal{N}(0, \sigma^2), 0]^T
\end{equation*} 
where $\mathbf{y}$ is the output (position), $\mathbf{y}^{m}$ is the measured output (i.e. noisy) and $\sigma^2$ is the variance of the noise. We vary the standard deviation $\sigma$ in Fig. \ref{fig_simcom_noise}.

\begin{figure}
     \centering
		\includegraphics[width= 0.49\textwidth, trim={0cm 0cm 0 1cm},clip]{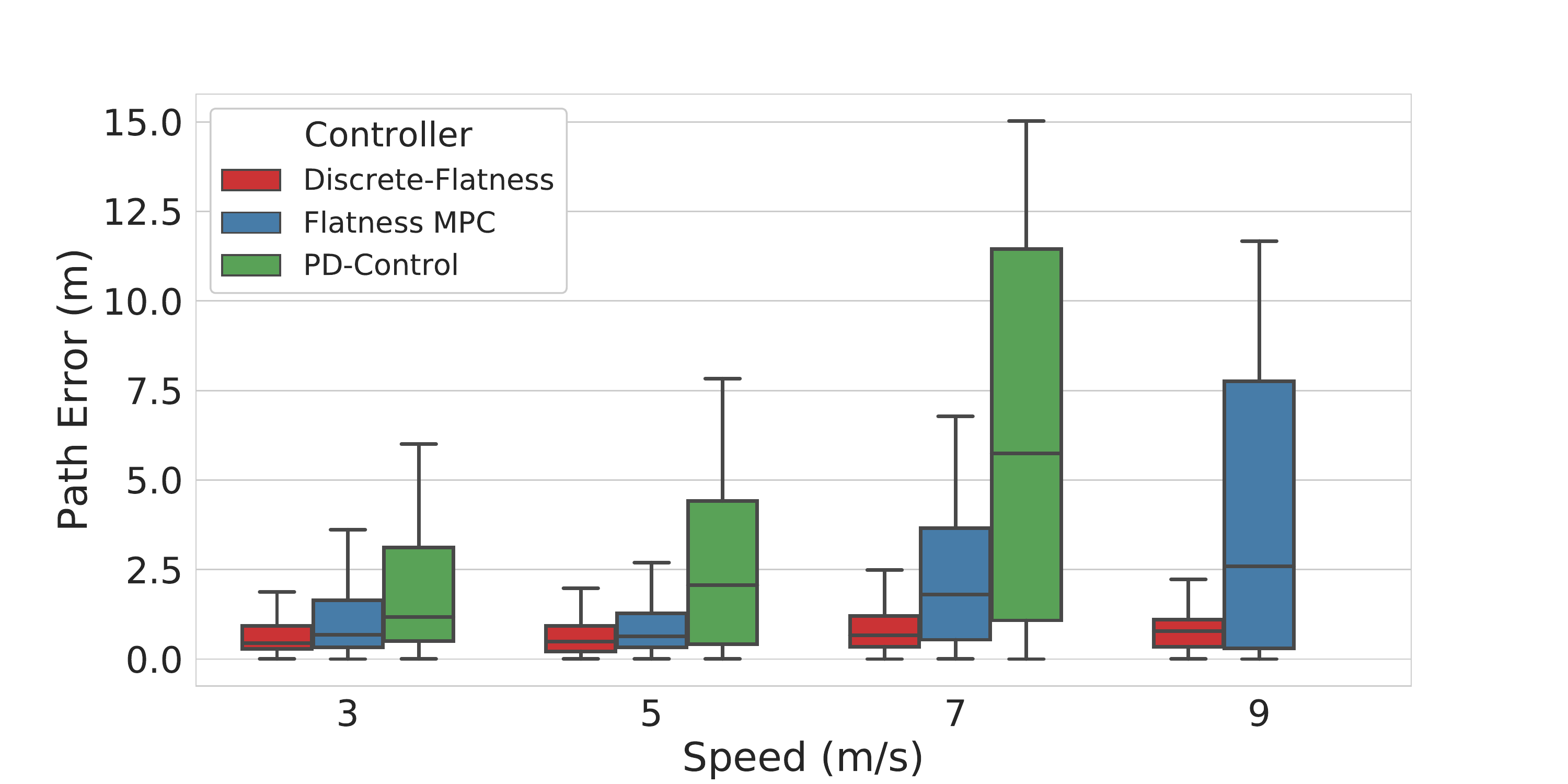}
	      \caption{Comparison of path error at increasing desired speeds for autonomous vision-based flight using \textit{PD Control} (green), \textit{Flatness MPC} (blue) and our proposed \textit{Discrete-Flatness} controller (red). The box plots are computed over 3 trials (for each controller at each speed) repeating the L-shaped path in Fig. \ref{fig_control_compare}. Our proposed approach \textit{Discrete-Flatness} (red) outperforms related controllers by performing prediction and not relying on noisy and delayed state estimation. The vehicle under \textit{PD Control} (green) goes unstable at 9 m/s. The performance under \textit{Flatness MPC} (blue) degrades as the speed increases. Our proposed \textit{Discrete-Flatness} (red) maintains low path error for all desired speeds.}
	      \label{fig_boxplot_control_compare}
	      \vspace{-0.6cm}
\end{figure}

\paragraph{Controller Comparison} We compare the effect of noise on the performance of \textit{Flatness MPC} and the proposed \textit{Discrete-Flatness} control. At each noise level or $\sigma$, we compute the average output error over $10$ trials for each controller. In Fig. \ref{fig_simcom_noise}, we observe that at low output noise (i.e. $\sigma < 0.0005$) both controllers have similar performance. However, as the output noise increases the performance of \textit{Flatness MPC} (blue in Fig. \ref{fig_simcom_noise}) significantly worsens as a result of heightened noise in the higher-order derivatives that comprise the flat state. Generally, this requires the design of an additional filter to reduce the noise in these derivatives. However, the filter introduces additional delays. In Fig. \ref{fig_simcom_noise}, we observe that working directly with the outputs (i.e., bypassing the state) in \textit{Discrete-Flatness} (red in Fig. \ref{fig_simcom_noise}) can result in improved performance robustness to output noise. 

\section{EXPERIMENTS}
\label{sec_exp}

\subsection{Hardware Set-up}

As shown in Fig. 1, we use a DJI Matrice 600 Pro with attached Ronin-MX gimbal. This system has a take-off weight of approximately 15 kg, and maximum span rotor-tip-to-tip of 1.64 m. All processing, including visual navigation, planning and control happens on-board the UAV on the primary computer, NVIDIA Tegra TX2 module (6 ARM cores + 256 core Pascal GPU). The primary computer uses a serial Transistor-Transistor Logic (TTL) connection to connect with the on-board M600 autopilot which provides vehicle data (gimbal state, autopilot state, e.t.c.) and the interface for sending control commands to the vehicle. We use a StereoLabs ZED stereo camera mounted to the Ronin-MX gimbal to provide greyscale imagery with resolution 672 x 376 at 15 Hz. The Tegra TX2 runs NVIDIA L4T v28.2 (a variant of Ubuntu 16.04 for ARM architectures). An XBee 900 Mhz wireless link is used to communicate with a ground station where we can monitor any status changes and send high-level commands (i.e., changing autonomy states). For more details, see \cite{c20}. 
\begin{figure}
     \centering
	\includegraphics[width=0.35\textwidth, trim={0cm 0.8cm 1.3cm 1.6cm},clip]{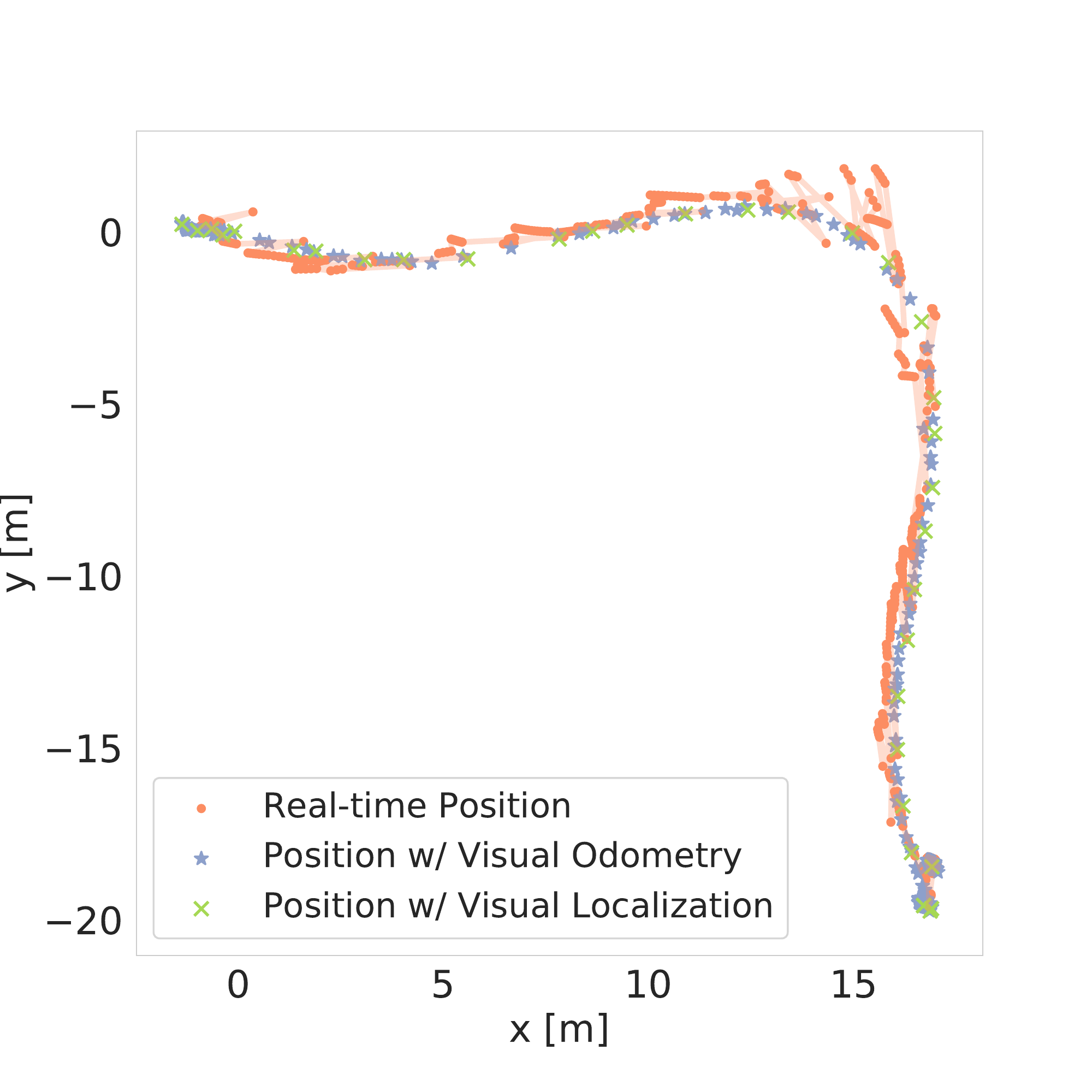}
	\caption{Comparison of the estimated noisy real-time position at 50 Hz (red), the estimated position from VO and bundle adjustment (blue) and including visual localization (green) for a path flown at 9 m/s. }
	\label{fig_pos_noise}
	\vspace{-0.7cm}
\end{figure}

\begin{figure*}
\centering
\begin{annotatedFigure}{
  \includegraphics[width=0.48\textwidth, trim={0cm 5cm 0 5cm},clip]{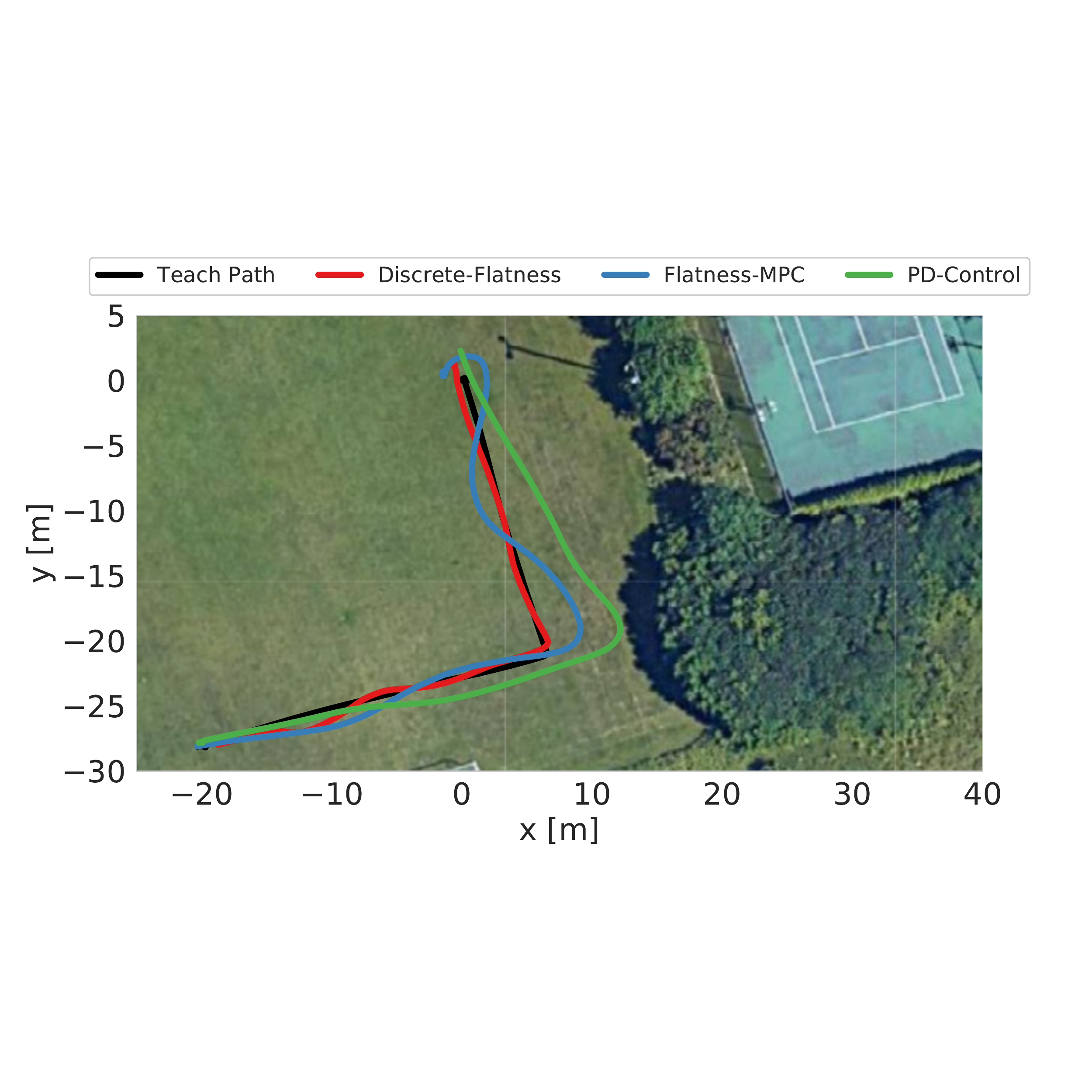}}
  \annotatedFigureBox{0.084,0.614}{0.084,0.614}{3 m/s}{0.8,0.314}\end{annotatedFigure}
 \vspace{-0.25cm}
\begin{annotatedFigure}{
  \includegraphics[width=0.48\textwidth, trim={0cm 5cm 0 5cm},clip]{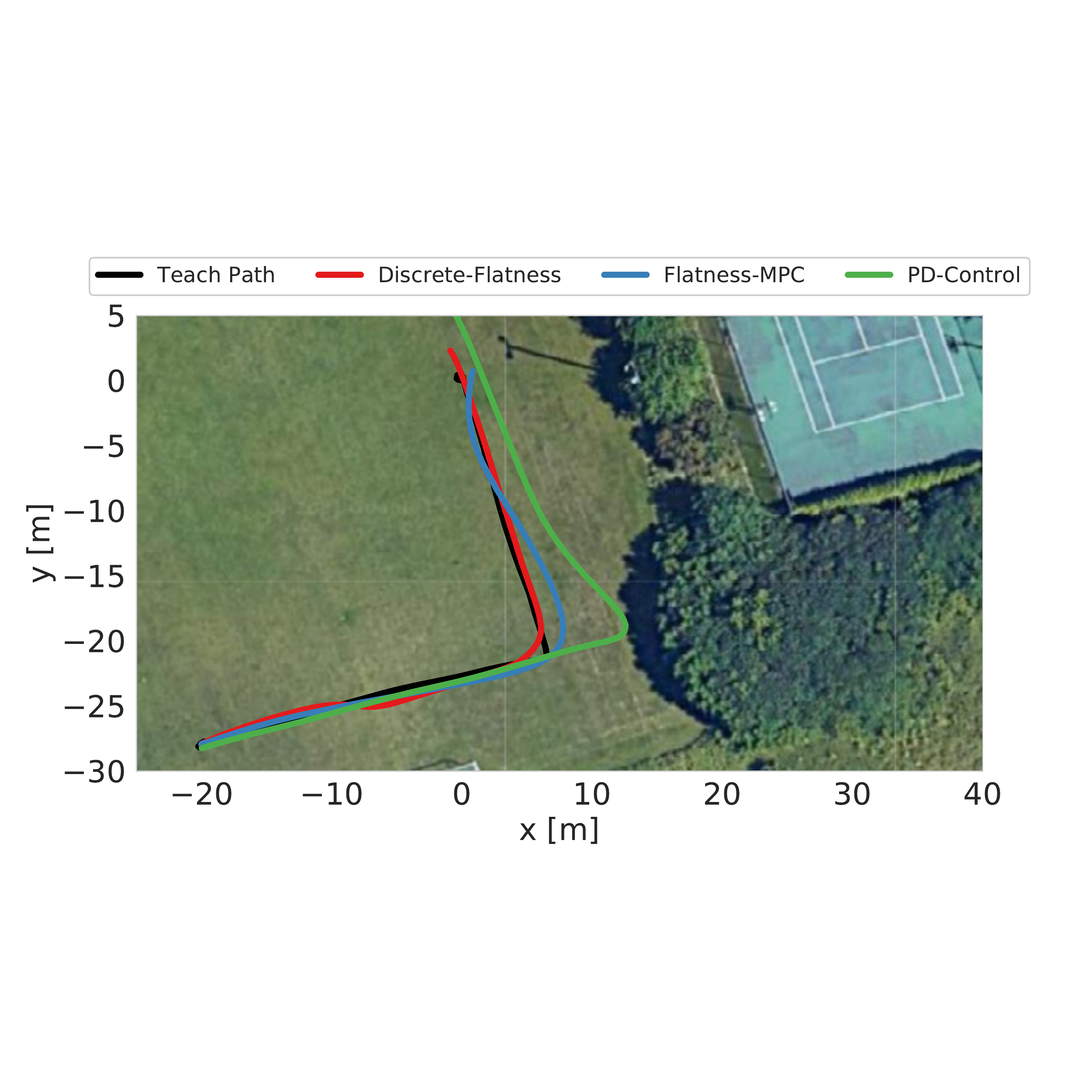}}
  \annotatedFigureBox{0.084,0.614}{0.084,0.614}{5 m/s}{0.8,0.314}\end{annotatedFigure}
 \vspace{-0.6cm}
\begin{annotatedFigure}{
  \includegraphics[width=0.48\textwidth, trim={0cm 5cm 0 7cm},clip]{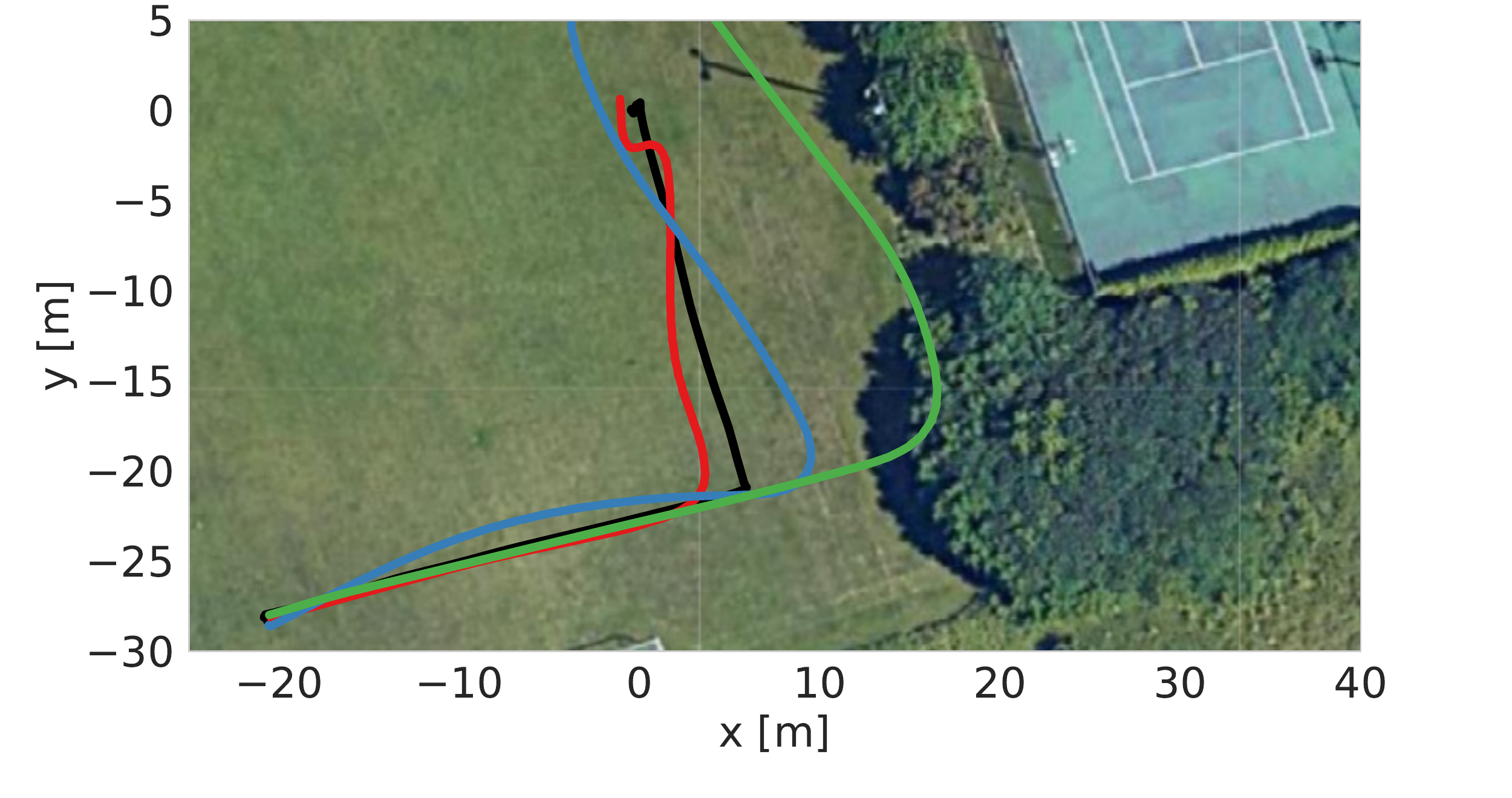}}
  \annotatedFigureBox{0.084,0.614}{0.084,0.614}{7 m/s}{0.8,0.354}\end{annotatedFigure}
\begin{annotatedFigure}{
  \includegraphics[width=0.48\textwidth, trim={0cm 5cm 0 7cm},clip]{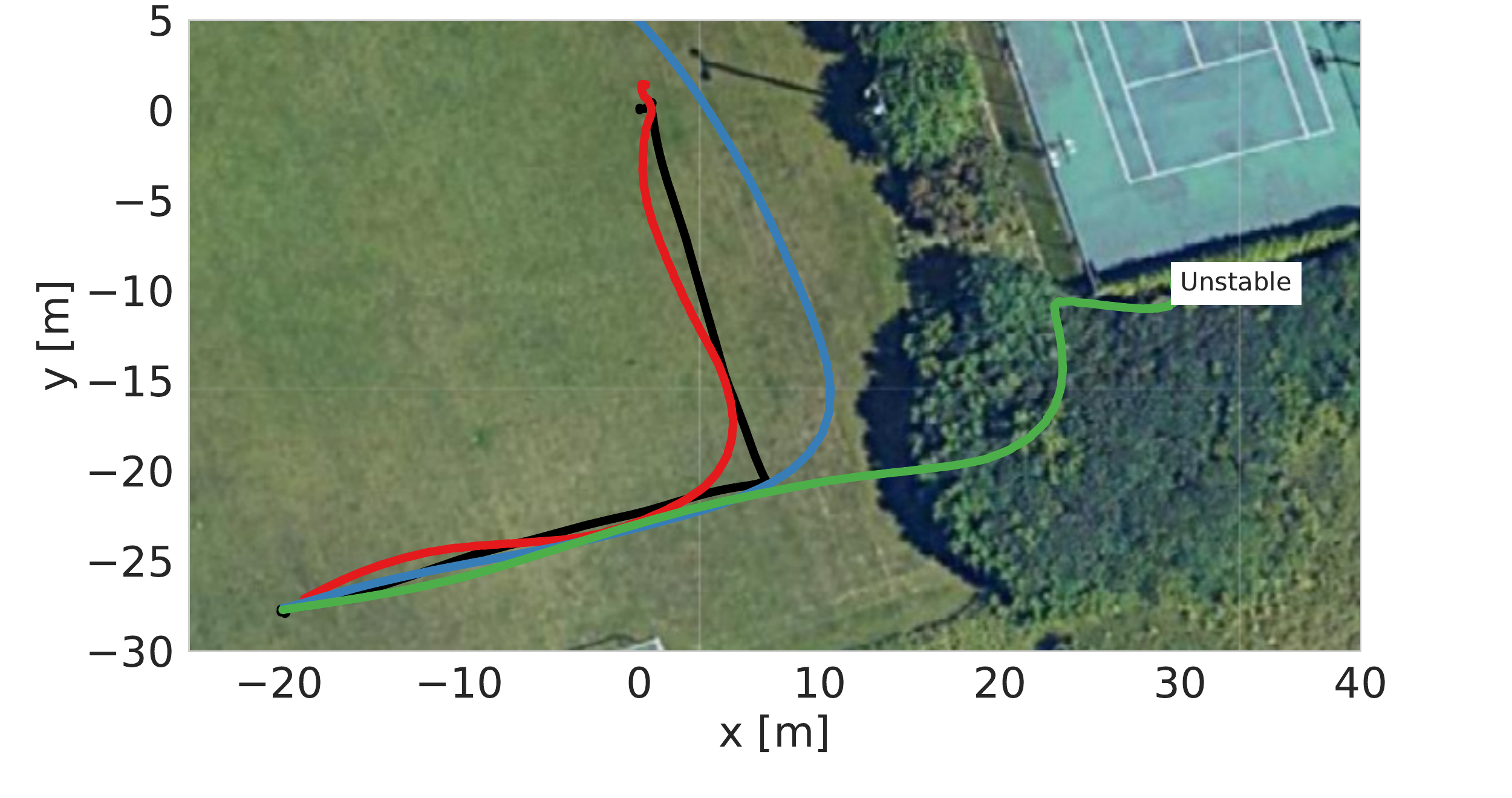}}
  \annotatedFigureBox{0.084,0.614}{0.084,0.614}{9 m/s}{0.8,0.354}\end{annotatedFigure}
\caption{Visualization of path flown when repeating the \textit{teach} path (black) under vision-based navigation using different controllers with a desired speed of (a) 3 m/s (b) 5 m/s (c) 7 m/s and (d) 9 m/s. The L-path starts at the lower left corner (at approximately $(-20, -30)$) and ends at the origin. We show one trial (out of three used in Fig. \ref{fig_boxplot_control_compare}) for each controller. We compare the paths flown using \textit{PD Control} (green), \textit{Flatness MPC} (blue) and our proposed \textit{Discrete-Flatness} (red). Our proposed \textit{Discrete-Flatness} (red) outperforms the alternative controllers that rely on noisy/delayed state estimates. Unlike the alternative controllers our proposed approach achieves high performance (see Fig. \ref{fig_boxplot_control_compare}) by turning before the corner even at high speeds.}
 \label{fig_control_compare}
 \vspace{-0.5cm}
\end{figure*}

\subsection{Visual Teach and Repeat Navigation Overview}

This section summarizes the visual teach and repeat (VT\&R) navigation system used in the experiments. In VT\&R, we consider two phases. During the \textit{teach} phase, the UAV flies using autonomous GPS waypoint following. The purpose of the teach path is to create a map of visual landmarks along the path. This taught path is stored as a set of vertices and edges which include the landmarks observed at each vertex. During the \textit{repeat} phase, GPS is disabled and the vehicle needs to perform autonomous navigation (of the teach path) while the VT\&R algorithm performs visual localization using a local segment from the taught path. In these experiments, we consider the \textit{controller implemented during repeat}.
\subsubsection{Teach}

During \textit{teach} only visual odometry (VO) is performed to generate a map. It does this by capturing image pairs from the calibrated stereo camera, and performing Speeded-Up Robust Features (SURF) feature matching with respect to the last dropped keyframe \cite{c20}. The raw matches are then passed through a Maximum Likelihood Estimation SAmple Consensus (MLESAC) robust estimator to find the relative transform to the last keyframe. If the motion exceeds a defined threshold, the current frame is set as a keyframe and the features (including the 3D landmark position associated with each feature) and vehicle-to-sensor transform are stored in a vertex in the pose graph. The relative transform is stored as an edge to the previous vertex. Windowed bundle adjustment (also called windowed refinement) is performed on the set of vertices. This set of dead-reckoned linked poses (after VO and bundle adjustment) represents the taught path to be repeated by our controller.

\subsubsection{Repeat}

During \textit{repeat} visual odometry (VO) is performed in the same manner as during teach with an additional thread running visual localization, i.e., visual matching to the map created during \textit{teach}, see \cite{c20} for more details. The high-rate real-time pose (or higher-dimensional state) used by the controllers is determined by extrapolating the last VO pose forward using Simultaneous Trajectory Estimation And Mapping (STEAM). STEAM was first introduced in \cite{c21} and details on the implementation used in this paper can be found in \cite{c22}-\cite{c23}. Noise in the high-rate real-time pose (and higher-dimensional state), see Fig. \ref{fig_pos_noise}, can come from many sources including stochasticity in the feature extraction and matching (both during VO and localization), any variable system delays (e.g., a delay in the gimbal state used to compute the vehicle-to-sensor transform), and the extrapolation using STEAM.   

\subsection{Comparison with Related State-Feedback Controllers}

We compare three controllers (\textit{PD Control} \cite{c20}, \textit{Flatness MPC} \cite{c15}\cite{c24}, and our proposed \textit{Discrete-Flatness} controller) implemented at 50 Hz. We consider only flight in 2-D (fixed yaw and no motion in the z-direction). All controllers are used to determine roll and pitch commands as described in Sec. \ref{meth_b}. 

\paragraph{Reference Generation} We use a simple reference generation approach. The geometric teach path is created by connecting keyframes from teach with straight lines segments. The reference position and velocity at time step $k$ for the non-predictive controller (\textit{PD Control}) is determined by finding the closest point on the path, i.e., $\mathbf{y}_{ref,k} = \mathbf{y}_{closest}$, and computing the reference velocity $\mathbf{v}_{ref,k}$ as the desired speed in the direction of the next keyframe. \textit{PD Control} \cite{c20} determines the commands at time step $k$ by weighting the position error $\mathbf{y}_k -\mathbf{y}_{ref,k}$ with the velocity error $\mathbf{\dot{y}}_k -\mathbf{v}_{ref,k}$. In the predictive controllers (\textit{Flatness MPC} and \textit{Discrete-Flatness}) we compute the reference on the path with a fixed desired speed. At time step $k$ we compute the reference by finding the closest point $\mathbf{y}_{closest}$ on the path, i.e., $\mathbf{y}_{ref,k} = \mathbf{y}_{closest}$. We compute the reference at the next time step by moving in the direction of the reference velocity (desired speed in the direction of the next keyframe) $\mathbf{v}_{ref,k}$ as $\mathbf{y}_{ref,k+1} = \mathbf{y}_{ref,k} + \delta t \mathbf{v}_{ref,k}$, for the prediction horizon $k+1, k+2, ... k+N$. The reference velocity $\mathbf{v}_{ref,k}$ changes as the output reference moves to the next straight line segment of the path. 
\paragraph{Parameters} We consider a lookahead of $2 s$ for \textit{Discrete-Flatness} and $1.5 s$ for \textit{Flatness MPC}. These were the maximum horizons for each controller that we could reliably compute at a rate of 50~Hz (20 ms). The lookahead difference is a result of \textit{Flatness MPC} requiring a few additional matrix multiplications at each time step. We tune the weights for all controllers to maximize stable performance. 
\paragraph{Results} We consider a simple L-shape teach path, black in Fig. \ref{fig_control_compare}, at a fixed altitude of 10 m above ground. In Fig. \ref{fig_boxplot_control_compare}, we present a box plot of the path error for each controller with desired speeds of 3 m/s, 5m/s, 7m/s and 9 m/s. These results are obtained from 3 repeated trials for each controller at each speed. As expected, the \textit{PD Control} (green) has the worst performance as it is unable to predict and react to the sharp turn in the path. At 9 m/s the \textit{PD Control} is not reliable and causes the vehicle to go unstable. \textit{Flatness MPC} (blue) has good performance for 3 m/s and 5 m/s, however, the performance dramatically degrades at 7 m/s and 9 m/s. Similar to the simulation in Sec. \ref{sec_sim}, \textit{Flatness MPC }(blue) relies on higher-order derivatives of position (velocity and acceleration). STEAM filters some of the noise in these estimates but introduces a delay. Consequently, inaccurate real-time estimation of these quantities prevents the \textit{Flatness MPC} from making an accurate forward model prediction. As such the vehicle does not turn before the corner (bottom right of black teach path in Fig. \ref{fig_control_compare}) for 7 m/s and 9 m/s. Our proposed \textit{Discrete-Flatness} (red) approach maintains a low tracking error at all speeds by turning before the corner as highlighted in Fig. \ref{fig_control_compare}.  

\begin{figure}
     \centering
		\includegraphics[width= 0.45\textwidth, trim={0cm 1.5cm 0 3cm},clip]{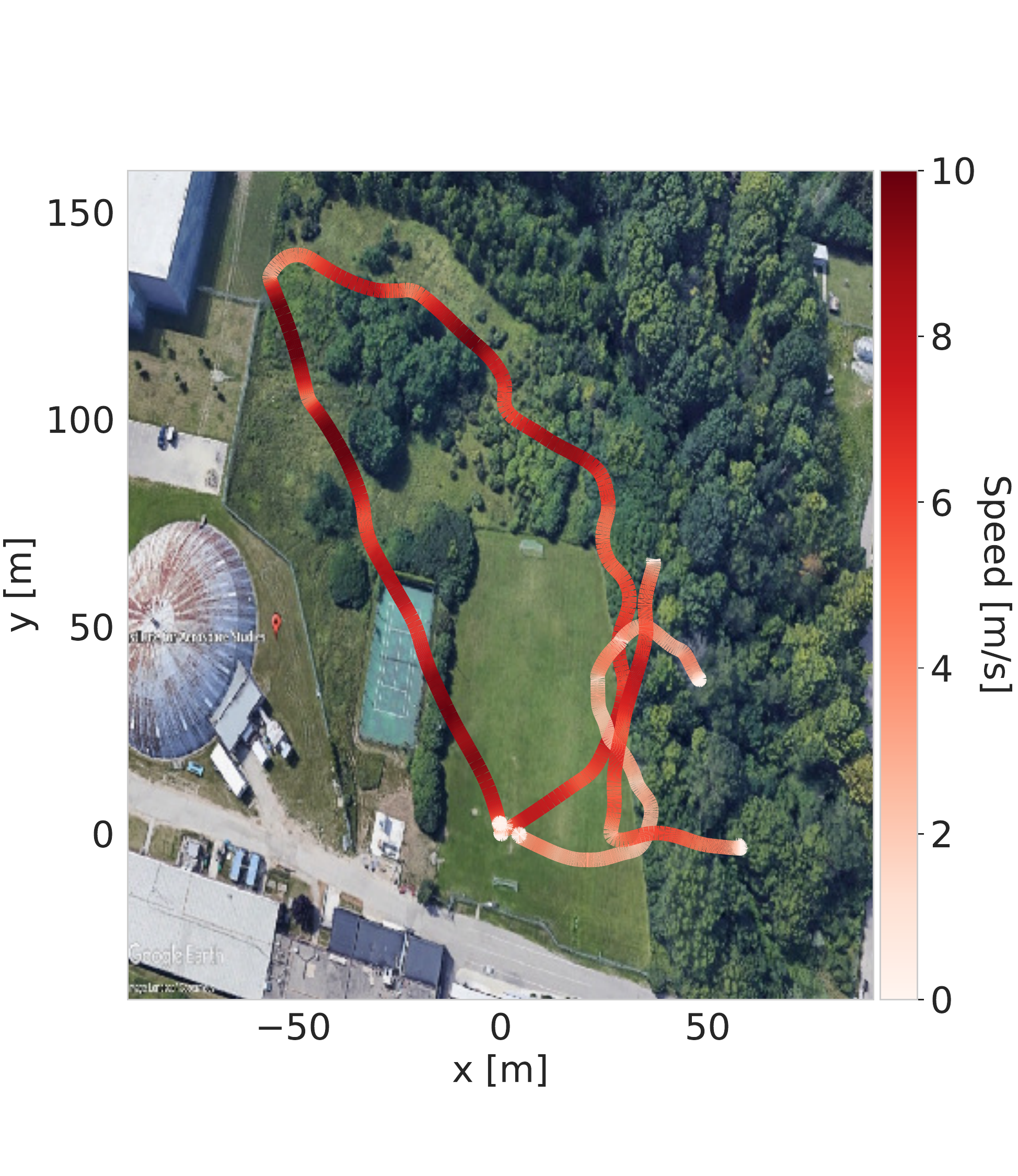}
	      \caption{Visualisation of 3 additional paths flown using vision-based navigation and the proposed \textit{Discrete-Flatness} control achieving speeds up to 10 m/s. We fly three paths: 1) a D-shaped path (at 10 m/s), 2) an S-shaped path (at 3 m/s) and 3) an L-shaped path (at 8 m/s). This spells out the abbreviation of our lab ``Dynamic System Lab" or DSL. }
	      \label{fig_dsl_visual}
	      \vspace{-0.6cm}
\end{figure}

\subsection{Demonstration of Discrete-Flatness up to 10 m/s}

We demonstrate the efficacy of our proposed \textit{Discrete-Flatness} controller for VT\&R at speeds up to 10 m/s by flying three paths (at fixed 20 m altitude above ground) - a D-shaped path (at 10 m/s), an S-shaped path (at 3 m/s) and an L-shaped path (at 8 m/s). The DSL flights are shown in Fig. \ref{fig_dsl_visual} with the speed profile overlay. Video: \url{https://tinyurl.com/flyoutthewindow}.

\section{DISCUSSION AND FUTURE WORK}
Exploiting discrete-time flatness for outdoor high-speed vision-based navigation, with potentially noisy high-rate real-time output (position) measurements, is a promising approach as it allows closed-loop visual autonomy without state estimation. Alternatively, the state estimate could be improved in the experiments by computing it using (\ref{eq_df_state}) from the window of inputs and flat outputs determined from discrete-time flatness. However, the proposed control architecture in Fig. \ref{fig_arch_diag} allows us to bypass this computation and use the outputs directly. This is particularly relevant to future work on learning-based control that could improve performance by simultaneously learning $F(\cdot)$ in (\ref{eq_outin}) and $F^{-1}(\cdot)$ in (\ref{eq_inout}) using \textit{only input and output data}. More traditional learning-based controllers require full-knowledge of the state to learn the dynamics model. Furthermore, this overcomes limitations when the state estimator and controller use different learned dynamics models. Another potential challenge is that the window size $r$ determined from first principles may differ for the actual system. In future work, we propose developing a learning-based controller by varying the window size and observing the effect on the learning performance. Future work is also required to extend the proposed control design to account for different input and output rates.

\end{document}